# Human-Vehicle Cooperation on Prediction-Level: Enhancing Automated Driving with Human Foresight

Chao Wang, Thomas H. Weisswange, Matti Krüger, Christiane B. Wiebel-Herboth
Honda Research Insitute Europe GmbH
Offenbach, Germany
{chao.wang@honda-ri.de, thomas.weisswange@honda-ri.de, matti.krueger@honda-ri.de, christiane.wiebel@honda-ri.de}

*Abstract*—To maximize safety and driving comfort, autonomous driving systems can benefit from implementing foresighted action choices that take different potential scenario developments into account. While artificial scene prediction methods are making fast progress, an attentive human driver may still be able to identify relevant contextual features which are not adequately considered by the system or for which the human driver may have a lack of trust into the system's capabilities to treat them appropriately. We implement an approach that lets a human driver quickly and intuitively supplement scene predictions to an autonomous driving system by gaze. We illustrate the feasibility of this approach in an existing autonomous driving system running a variety of scenarios in a simulator. Furthermore, a Graphical User Interface (GUI) was designed and integrated to enhance the trust and explainability of the system. The utilization of such cooperatively augmented scenario predictions has the potential to improve a system's foresighted driving abilities and make autonomous driving more trustable, comfortable and personalized.

*Keywords—Human-Machine Interface, Automated Vehicles, Novel Interfaces and Displays*

## I. INTRODUCTION

The development of automated driving (AD) aims at successively shifting control from the human driver to an autonomous system. However, as long as AD systems are still challenged by multiple factors [1], the human driver will have to stay in the "loop". Even in SAE Level 4/5 automation [2], where driving safety is to be ensured by the AD system, low efficiency and an uncomfortable driving experience can be the result of inappropriate behavior choices.

One common challenge for AD systems is the accurate and timely prediction of other road user's behavior. Human drivers are often better at integrating multiple information for estimating future events in complex or novel traffic situations. If, for example, a car in front just parked at the side of the road, a human driver may be anticipating someone to open a door and exit towards the road. For an AD system, it may just be a parked car. Such advanced reasoning becomes more and more difficult for any AD system with increasing dimensionality of the underlying feature space [3]. Another challenge of AD systems is to appropriately handle the user's trust in the system. Handing over control to an opaque AD system requires a substantial amount of trust, as the health of all vehicle occupants is at stake. Even a perfectly functioning AD system may thus not be utilized or at least create a feeling of discomfort for a long period until the system has proven itself trustworthy to the human driver.

One popular approach to cope with actual AD system limitations is to hand over full control back to the human driver until the difficult situation is resolved [4], [5]. Similarly, drivers may take back vehicle control whenever they do not trust the AD system to handle a situation well, regardless of actual AD capabilities. However, human driving behavior can be impaired after a take-over [6], for example due to a lack of situation awareness [7]. Therefore, human-vehicle cooperation has been suggested as a promising alternative concept [8][9][10]. Wang et al. [11], [12] have previously proposed a refined framework of human-vehicle cooperation which can guide concrete implementations of cooperative HMI's. The framework structures the interaction between human and an AD system in four levels: perception-level, prediction-level, plan-level and control-level. Most of the cooperation concepts between humans and automated systems have been suggested on the plan- [13][14][15]or control-level [8][16]. The most prominent approach of the former is to allow the human driver to intervene by selecting behaviors for the AD system ("shared control")[17]. However, the provided options for an alternative maneuver might not reflect the true reason for a human driver's wish to intervene and therefore can be inappropriate to solve the situation.

Recently, Wang et al. [12] suggested another cooperation approach in which the AD system can directly benefit from the human driver's assessment of the situation on the prediction-level, which then leads to an adjustment of the behavior planning. The concept of human prediction-level intervention was evaluated in a Wizard-of-Oz driving simulator study. Participants could point out potentially dangerous vehicles by looking at them and saying, "Watch Out". Following a standardized protocol, the experimenter would adjust the driving behavior of the simulated AD car accordingly (slow down or change lane). Results confirmed the perceived usefulness of the prediction-level intervention and the feasibility of conveying predictive information through gaze. However, participants also reported that the usability of the intervention function could be increased by making the systems' current maneuver plan and predictions of other traffic participant's behavior transparent to the user. Therefore, an important next step in order to investigate the potential of the proposed concept is to implement it within a real AD system. This allows for 1) a more thorough evaluation of the effects of the prediction-level intervention on the 'real' AD system behavior and its perceived usefulness and 2) the possibility to make the real system state and behavior planning transparent to the user by providing at the same time a direct feedback on the effects of the user's interventions.

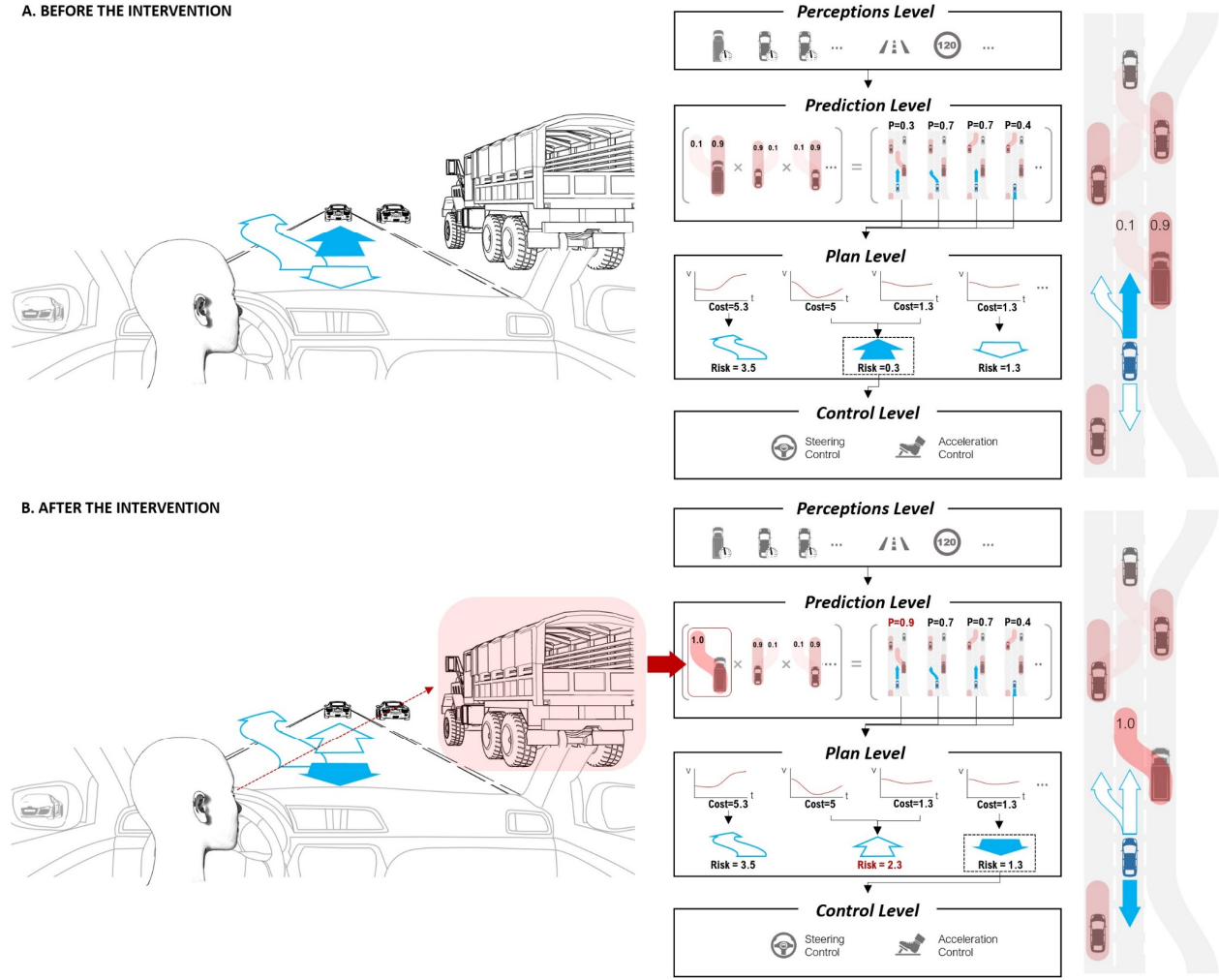

Figure 1 Processing flow of iTFA (A) and its adaption with human driver's intervention (B).

Here, we set out to realize a first, fully functional implementation of such a prediction level intervention function within an AD system. To that end, we use an AD system based on the "intelligent Traffic Flow Assistant" (iTFA) [18] which was previously demonstrated to work in public traffic and which features an advanced behavior prediction component for highway driving.

## II. CONCEPT, SYSTEM DESIGN AND IMPLEMENTATION

### A. Automated driving by iTFA and opportunities for cooperative enhancement

The AD system that we use is based on the "intelligent Traffic Flow Assistant" (iTFA) [18], which was previously used to demonstrate partially automated driving on public highways. Besides a variety of perception modules, it features a planning component that optimizes the future driving trajectory of the vehicle integrating safety and comfort considerations. Additionally, it includes a component for advanced predictions of the behavior of other traffic participants, which was previously also used to enhance classical ADAS applications [19]. This component analyzes the situational context of another vehicle to compute a probability for a cut-in maneuver on the highway. The models are based on specific situation motifs common to many real-world cut-in examples rooted in overtaking maneuvers, for example a faster car approaching a much slower car on the same lane while there is a sufficient gap on its neighboring lane. If the predictions are incorporated into the planning, the automated vehicle, for example, slows down or changes lane before another vehicle enters its lane and by that can prevent stronger decelerations which would happen with more reactive approaches. Cut-ins which are done for other reasons or that happen in "untypical" situations cannot be covered with this prediction (but also see [1]) and the automated vehicle will therefore only react after the other vehicle enters its lane. There are also additional reasons that make it difficult to predict every cut-in early enough, such as the limited range and possible occlusions of the sensors.

Cut-ins that are not covered by the prediction framework provide opportunities for a human driver to improve the performance of the automated driving system. However, there is no need for a full take-over of control as all the other parts of the systems are not affected. The concept presented in this paper allows a human to infuse an additional prediction while the automated car continues to take care selecting an appropriate maneuver.

### B. Integrating prediction-level intervention into iTFA system

In this section, we introduce the concept of prediction-level cooperation with an example and explain how a human driver's intervention is integrated into iTFA. Figure 1A shows the processing flow of iTFA. After the system perceives the environment through various sensors and recognizes different objects ("Perception Level"), it starts to predict the behavior of each relevant traffic participant ("Prediction Level" - left). The output of the iTFA prediction module is a list of possible future scene compositions incorporating the predicted behaviors of vehicles surrounding the automated car and the probability of each composition. Additionally, the probabilities are conditioned on different behaviors of the ego vehicle itself (driving straight, lane change left, lane change right) – a cut-in of a neighboring vehicle might be more likely, if its driver assumes that the ego-car is "giving way" through a lane change than if it is assumed that it continues driving in its lane. The final probabilities are a combination of the individual (conditioned) behavior predictions of the relevant cars ("Prediction Level" - right).

The planning component of iTFA takes the most likely compositions as input and performs an optimization on a parametrized representation of the future ego trajectory in terms of acceleration and lateral position over time ("Planning Level" - top). The optimization objective contains terms for safety (e.g., distance to other vehicles), utility (e.g., deviation from desired velocity) and comfort (e.g., punishing high jerk). The behavior selection component receives the optimized trajectories and their costs for each possible future composition and selects the trajectory that has the best combination of composition probability and cost ("Planning Level" - bottom) to be commanded to the "Control Level". A GUI provides the driver with information about the situation as it is perceived by iTFA, along with the currently planned trajectory and the predicted behaviors of the surrounding vehicles. Additionally, iTFA will activate the respective indicator signal 3 seconds before starting a lane change maneuver.

In the example situation shown in Figure 1B left, there is a truck on a merging lane in front of the ego vehicle. The iTFA system will predict that the truck will continue to drive straight with a high probability, as the situation does not fit any of its cut-in models. Based on this prediction the automated vehicle will decide to continue going straight with constant velocity. However, a human driver might have a different interpretation of the situation. He could conclude that the truck wants to change to the left lane as he saw it coming from the entrance ramp and getting closer to the end of the merging area. If this would happen the automated vehicle would be required to decelerate strongly once the truck enters its lane which could be dangerous and uncomfortable. Thus, the driver might want to proactively intervene by means of informing the system that the truck may cut in (Figure 1B). The system interprets the intervention as to increase the probability of the truck changing lane to 1.0. All other parts of the system are left untouched, so it will simply incorporate this prediction in both, computing future scene compositions and providing optimized trajectories for these scenes. In this way, the user does not have to think about the best reaction to a future cut-in, for example by checking if an own lane change would be possible but leave this to the automated vehicle. Therefore, the safety cannot be compromised by user intervention on the prediction level. In the example in Figure 1B, the injected prediction will result in a different behavior selection – the automated vehicle starts decelerating smoothly to predictively increase the gap for when the truck is cutting in. The prediction will stay at 100% until the scene changes significantly - either when the respective vehicle becomes irrelevant (e.g., after overtaking it) or when it finishes its lane change. At the same time, the automated vehicle will also continue to react to additional predictions that come from within the system.

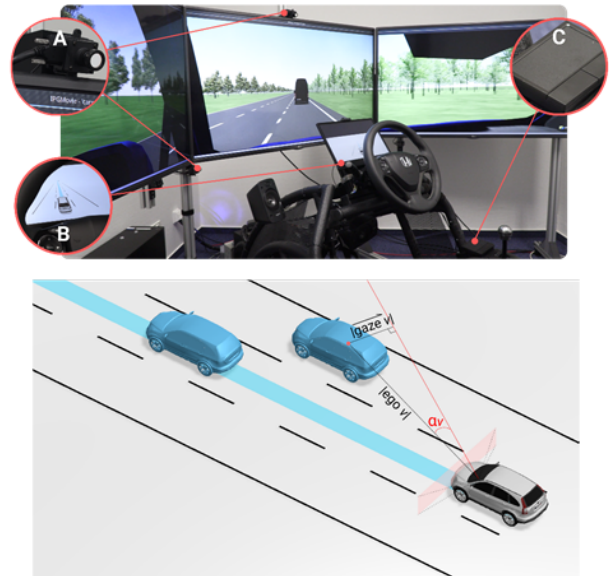

Figure 2 Top: The layout of the prototype in the driving simulator environment. Bottom: Gaze-vector and distance-based vehicle selection method

### C. Interface design and implementation

- Setup of the prototype

For testing the concept, a prototype was implemented in a driving simulation environment (IPG CarMaker [20]). Three display panels (50-inch diagonal, Resolution: 3 x 1080p, 60 Hz) were arranged to provide approximately 160-degree field of view of the driving scene rendered with IPG CarMaker 9.1. A remote eye-tracking system (Smart Eye Pro [21], see Figure 2 Top A) was used for gaze recording. A 14-inch screen was mounted in front of the steering wheel to display the GUI (Figure 2 Top B). Buttons were fixed to the console next to the driver seat for explicit driver input (Figure 2 Top C).

- Gaze-based object referencing

Prior studies suggested gesture-based interaction as a promising means for referring physical items around ego vehicle [22], however, it was also reported that gesture interaction increased the driver's workload [23], [24]. In the study by Wang et al [12], gaze- and speech-based interaction was validated. The result showed it was easy to refer to object in the environment via gaze, but confirmation through speech was not quick enough. Therefore, we use a button press to trigger a gaze-based vehicle reference in this study. This is obtained via remote eye-tracking which captures the driver's gaze point on the driving simulator displays. To map this 2D gaze point to the driving simulation environment, a 3D gaze vector is computed based on the virtual origin of the simulation world view and the screen intersection point in simulation coordinates (see Figure 2 Bottom).

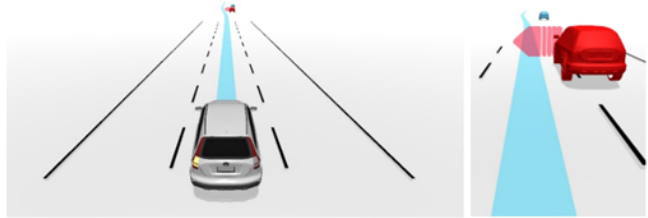

Figure 3 Left: the overview of the GUI interface; Right: if a vehicle is selected by gaze-button input, it turns red.

When looking at the screen representation of a vehicle, a virtual ray along the gaze vector should intersect with the corresponding vehicle in simulation space and thus provide the basis for gaze-based referencing. However, such an intersection is not just dependent on the gaze direction but also on the eye-tracking accuracy and the distance towards the respective vehicle because angular errors are amplified by distance. This makes a pure intersection-based selection impractical. To overcome these issues, we base the selection on an error measure ($e$) that is intersection-independent (see Eq(1) ). This procedure makes sure that a vehicle is selected if present in the driver's field of view.

In a first step, the perpendicular distance between the 3D gaze ray ($\overrightarrow{gaze}$) and each virtual vehicle ($v$) is calculated. To overcome the proximity bias, this distance is divided by the distance between the respective vehicle ($v$) and the ego-vehicle (ego). The resulting measure is the sine of the absolute deviation ($\alpha$) from the gaze angle towards the respective virtual vehicle $v$.

$$e_v = \sin \alpha_v = \frac{|\overrightarrow{gaze}\ v|}{|\text{ego}\ v|} \quad (1)$$

Accordingly, we define the selected vehicle as the vehicle that minimizes $e$ at the time of the button press.

$$v_{\text{select}} = \underset{v\ \in\ \text{traffic with}\ \alpha_v \in [0, \pi/2]}{\arg\min}\ e_v \quad (2)$$

For our present implementation we assume that a user will only want to inject predictions that are relevant for the current driving path. Therefore, we automatically determine the predicted lane change direction based on the relative position of the vehicle selected by gaze. This means selecting a vehicle on the right lane will communicate a "change left" prediction to the system and vice versa for left lane vehicles.

- GUI output

In the study by Wang et al. [12] users reported a wish for more transparency of the perception and planning of the AD system and the gaze-based vehicle selection, to enhance trust and explainability. Previous research suggested that showing the intent of an automated driving system through an HMI contributes to trust [25]. Therefore, we implemented an interface which visualizes the detection of surrounding vehicles, the ego vehicle's maneuver plan and available predictive information (Figure 3). Four layers of information are shown parallel:

1. Road information: lane marks detected by the vehicle are displayed dynamically to enable the driver to know the current lane of the ego vehicle,

2. Ego vehicle: A 3D model of the ego vehicle is shown in the center of the screen from a third-person perspective. The braking and indicator lights mimic those of the actual ego vehicle,

3. Ego-vehicle maneuver plan: a planned driving trajectory is shown as an arrow on the road, and

4. Traffic information and behavior predictions: The 3D models of other vehicles detected by the system are shown.

If the iTFA system judges the probability for another vehicle to change lanes to be above a threshold value, a red arrow will be shown on the side of the respective 3D model. Models of the vehicles selected by the driver by gaze-button input change their color to red, thus informing the driver about the recognition of his input.

III. SCENARIOS, TEST DRIVING AND RESEARCH HYPOTHESES

We created a number of highway scenarios that can be used to test the interaction with the system and will be the core part of a future user study. In all scenarios, the driver is free to inject information about any other traffic participant at any time. We introduce each scenario together with a description of a test drive that we carried out.

In the first scenario, the ego vehicle is driving on the center lane, approaching a car on the right lane that is closing in on a slow truck. During test driving, when getting closer to the target car, the automated vehicle performed a lane changes well ahead of the start of the cut-in maneuver of the other vehicle, as this is a case explicitly modelled in iTFA. As the prediction and future plan of the system were visualized to the driver there was no need for the driver to inject additional information in our test drive.

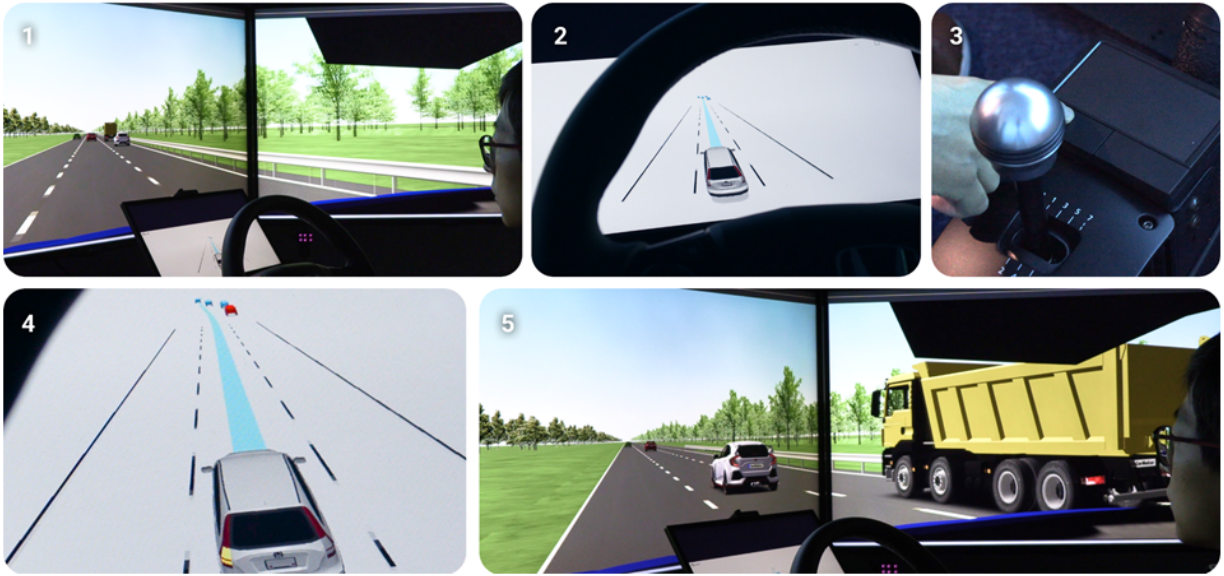

Figure 4 The interaction flow of the second scenario: 1) A sports car is driving behind a truck. 2) The system does not predict the car to change its lane 3) The driver clicks the button while gazing on the sports car indicating that it may change lane. 4) After the system has received the driver's input: a sound feedback is provided; the sports car is highlighted in red in the GUI; iTFA changes the ego vehicle's planned trajectory to a lane change. 5) The system maneuvers the vehicle accordingly.

In the second scenario, there is a sports car driving behind a slow truck on the right lane (Figure 4). In our test drive, the human driver detected the sports car and, as he inferred that it would overtake the truck as soon as there is a gap, he selected the vehicle by gaze and injected a prediction into the system. At that time, iTFA planned to continue driving straight as it did not have an indication for the sport car to change its lane. However, after receiving the human prediction, the system initiated a lane change, and when the sports car started its lane change the ego vehicle was already moving and did not need to decelerate.

In the third scenario, the ego-vehicle is approaching an entrance lane with a van that is about to enter the highway. For the human driver in the test run, it was easy to predict that the van would have to change lanes before reaching the end of the entrance ramp. As the GUI showed no sign that the automated vehicle was aware of this, the driver communicated his prediction to the system. In response to this, the ego-vehicle was slowing down smoothly to let the van enter, as another vehicle on the next left lane was preventing a lane change to the left. In contrast, when the same situation was driven without human support, the ego-vehicle was braking strongly when the van entered the highway, as iTFA did not have means to anticipate the cut-in early enough. The interaction between the system and the driver worked smoothly in our test drives and we did not experience any unexpected behaviors. We therefore conclude that the implementation is fully functioning and can now be used for further investigations.

We noticed that the time when both human and system predictions become evident is only a few seconds, so the driver was occasionally injecting a prediction although the system was already aware of it. It will therefore be interesting to investigate how this might influence user ratings in our upcoming study.

In a next step, we plan to conduct a user study which should evaluate, if the proposed concept of prediction-level cooperation does improve the driving experience and helps users to cope with low trust. This will involve both, objective data, for example comparing the vehicle deceleration values between automated and cooperative settings, and subjective measures, such as user experience ratings. One hypothesis is that participants would still subjectively perceive the driving experience as more comfortable even if their input does not significantly change the vehicles trajectory, due to an increased feeling of control. Besides, the timing, location and consequence of the input can be different form person to person. For example, not everybody would point out the sports car to change lane. Another interesting question is thus to investigate the relationship between different participants' interventions and user experiences.

## IV. CONCLUSION

In this work, we realized a concept of prediction-level intervention with a real AD system. In addition, prior feedback from participants suggested a lack of means to develop sufficient system understanding and to support appropriate trust calibration. Therefore, here a graphical user interface was designed to visualize the influence of the user's input on the prediction, plan and control process of the AD system. We implemented a functional prototype in a driving simulator environment and created three typical driving scenarios in which an AD could benefit from a human driver's anticipation capabilities to improve driving efficiency, comfort and trust handling without compromising safety. Whether these theoretical advantages also emerge in real use will be subject to future investigation. As the automated driving framework has previously been demonstrated on public roads, validation of the concept in real world scenarios will also be feasible after the simulator study.

## V. ACKNOWLEDGEMENT